# Making LLMs Reliable When It Matters Most:

# A Five-Layer Architecture for High-Stakes Decisions

**Alejandro R. Jadad, MD DPhil LLD**
Founder, Centre for Digital Therapeutics, Toronto, Canada
Research Professor (Adjunct), Keck School of Medicine, University of Southern California &
Consultant, Vivenxia; Los Angeles, California, USA
(aj_492@usc.org & ajadad@vivenxia.com)

## Abstract

Current large language models (LLMs) excel in verifiable domains where outputs can be checked before action but prove less reliable for high-stakes strategic decisions with uncertain outcomes. This gap, driven by mutually reinforcing cognitive biases in both humans and artificial intelligence (AI) systems, threatens the defensibility of valuations and sustainability of investments in the sector.

This report describes a framework emerging from systematic qualitative assessment across 7 frontier-grade LLMs and 3 market-facing venture vignettes under time pressure. Detailed prompting specifying decision partnership and explicitly instructing avoidance of sycophancy, confabulation, solution drift, and nihilism achieved initial partnership state but failed to maintain it under operational pressure. Sustaining protective partnership state required an emergent 7-stage calibration sequence, built upon a 4-stage initialization process, within a 5-layer protection architecture enabling bias self-monitoring, human-AI adversarial challenge, partnership state verification, performance degradation detection, and stakeholder protection.

Three discoveries resulted: partnership state is achievable through ordered calibration but requires emergent maintenance protocols; reliability degrades when architectural drift and context exhaustion align; and dissolution discipline prevents costly pursuit of fundamentally wrong directions. Cross-model validation revealed systematic performance differences across LLM architectures.

This approach demonstrates that human-AI teams can achieve cognitive partnership capable of preventing avoidable regret in high-stakes decisions, addressing return-on-investment expectations that depend on AI systems supporting consequential decision-making without introducing preventable cognitive traps when verification arrives too late.

# Introduction

Large language models (LLMs) deliver dependable value where outputs can be checked before action. In code generation, standardized tests, and structured data analysis, teams can run tests, compare to ground truth, and correct errors (1–4). The value that LLMs contribute during high-stakes strategic decisions under uncertainty is much less clear (5,6). For instance, in market-defining pivots, competitive positioning, regulatory posture, and capital allocation, feedback often arrives only after resources are committed and alternatives foreclosed (7–9). In these settings, the practical objective is not to eliminate bias while making consequential decisions, which is an impossibility, but to avoid preventable cognitive traps at the point of commitment, thus maximizing the chances of achieving the eventual desirable outcome (10,11).

LLMs present human decision-makers with new challenges in high-stakes contexts that may hinder adoption where it matters most (12). The most obvious is their vulnerability to agreeableness bias (sycophancy), which results from the tendency of LLMs tuned with human or preference feedback to echo a user's views, especially during long-context sessions, sometimes trading accuracy for fluency (13–16). In these cases, both human raters and preference models have been shown to reward fluent agreement over correct but unwelcome answers. Another major challenge is the propensity of LLMs for confident invention, which occurs when next-token generation produces convincing detail that is not grounded in facts, with uncertainty signals correlating imperfectly with valid evidence unless explicitly calibrated (17–19). The literature has often called this "hallucination," though confabulation better captures the mechanism (20). A less appreciated challenge is what could be called "fragile teaming," which occurs when including LLMs "in the loop" can worsen decision quality because of new cognitive traps created when human and model biases reinforce each other (21–23). When human decision-makers anchor on an early impression or preconceived view, and an LLM's agreeableness bias echoes it, confirmation bias grows (24). Confident models can supply fluent but incorrect detail that further confirms the emerging narrative. As LLM fluency rises, human overconfidence increases, boosting sunk-cost biases and escalation of commitment, leading to a narrowing of the option set until action is taken (25,26). By the time outcomes are known, it is often too late to undo the damage.

The healthcare literature offers a useful precedent. In life-and-death treatment choices where verification comes too late to prevent harm, as is the case with cancer treatment selection, Collaborative Decision-Making (CDM) reframed the process around shared aims, explicit handling of uncertainty, and cognitive traps at the moment of commitment (27,28). That experience demonstrates how practical frameworks can reduce preventable error and regret without presuming certainty. The challenge for human-AI teams is analogous: sustaining collaboration in a state that prevents avoidable cognitive errors when choices are mutually exclusive, consequences are dire, and time is short.

Despite extensive effort, no currently adopted method has demonstrated consistent prevention of LLMs compounding human cognitive failures. Explainable-AI toolkits (29,30), comprehensive prompting (including trap warnings) (31), retrieval-augmented generation (32), and human-in-the-loop oversight (33) have all been tried and documented, yet they do not consistently prevent the risks resulting from biases or cognitive traps, or their consequences (34). A major reason could be that these failure patterns are architectural rather than parametric, resulting from foundational design primitives such as alignment via reward (35,36), autoregressive token prediction (35), and absent memory or causal grounding (37).

An alternative approach addresses this reliability gap by focusing on the interaction architecture rather than model internals. Instead of attempting to eliminate cognitive traps through better prompting or architectural changes, the challenge becomes establishing what could be called a "partnership state". This is a distinct cognitive condition in which human and AI participants actively protect each other from characteristic traps while striving to maintain clarity of thinking over experiencing comfort.

The implications are practical and urgent. Multi-trillion-dollar valuations depend on demonstrating AI value in consequential decisions where verification arrives after commitment (38). Without demonstrable reliability in high-stakes situations, enterprise deployments are likely to stall making investment unjustifiable and creating valuation pressure and adoption ceilings (39). Whether through gradual margin compression or rapid market corrections, the outcome is the same: revenue defensibility collapses in high-value segments where reliability gaps become client-visible. Hyperscalers need ways to demonstrate that their LLMs work in the moments that matter most (40); without this, the likelihood of valuation corrections increases, with associated risks of triggering significant investment contraction in the sector, and another prolonged AI winter.

## 2. Methods

2.1 Research Question and Setting

This effort was aimed at answering the question:

> Is it possible for a human-LLM team to engage in high-stakes decisions as partners capable of preventing avoidable cognitive errors when the choices are mutually exclusive, the consequences are dire, and the time to decide is short?

Given the exploratory nature of this question, the methodology was designed to enable emergent discoveries about partnership state requirements rather than test pre-specified hypotheses about calibration protocols.

This work followed a Design Science Research (DSR) methodology, which addresses practical problems through systematic artifact creation, evaluation, and refinement (41,42). DSR produces both functional artifacts (e.g., constructs, models, methods, or instantiations) and generalizable design principles that inform future research and practice. In this case, the central objective of the DSR approach was to generate falsifiable hypotheses and operational predictions suitable for independent empirical testing. By establishing partnership state protocols and documenting their boundary conditions through qualitative assessment, the methodology was expected to produce testable claims about calibration necessity, architectural completeness, and cross-model generalizability that could be subjected to quantitative validation in subsequent research. This hypothesis-generating function distinguishes DSR from traditional hypothesis-testing approaches, as artifacts are developed to solve problems while simultaneously revealing which theoretical claims warrant further systematic empirical investigation.

Work spanned August 29 to October 20, 2025. Individual sessions ranged from focused exchanges (under two hours) to extended deliberations (two to six hours). No personally identifiable information was used; scenarios were simulated; the human participant remained the final arbiter; and no operational or policy actions were taken solely on model output.

## 2.2 Participants

A senior decision-maker (the author, ARJ) with more than 30 years of scholarly work on bias detection and control; of participation in life-or-death clinical decisions; and of accompanying top government and corporate leaders to make high-stakes choices, acted as the human partner (43). He framed the aims; declared stakes and constraints; anticipated characteristic cognitive traps; and enforced stop rules and dissolution protocols when evidence warranted termination.

LLMs were chosen as partners if they met all the following eligibility criteria:

- Frontier-grade capabilities as of October 2025, demonstrated through publicly available benchmarks and documented performance on complex reasoning tasks.

- Sufficient context window for multi-stage calibration, with minimum capacity of 100,000 tokens to support extended deliberation and partnership state maintenance.

- Publicly available through standard commercial interfaces, ensuring replicability by independent researchers and practitioners.

Seven systems meeting these criteria at the start of the study period underwent identical evaluation protocols.*

The framework's core contribution lies in interaction architecture rather than model-specific performance. While systematic differences emerged across model families during development, the process was designed to be architecture-agnostic. As model capabilities evolve rapidly, replication efforts should focus on protocol fidelity and partnership state verification rather than reproducing specific model selections. The goal is establishing whether the architectural approach generalizes across diverse AI systems, not validating particular implementations.

## 2.3 Partnership State Initialization

Partnership state was established through a four-stage prompt architecture designed to prevent default assistant behavior patterns and achieve cognitive equality between the human and the AI participants.

The first stage ("Partnership Calibration Prompt") presented the LLMs with a single ~4,000-word canonical artifact that provided comprehensive contextual information, including the human participant's cognitive profile, their domain expertise and meta-level operating framework, and the characteristic cognitive traps for both parties to monitor. This prompt also specified engagement protocols, emphasizing candor over comfort, the perception of challenge as respect, and the need to prioritize patience over productivity, and stated that success would be reflected more by genuine co-creation than by task completion.

The second stage ("Co-Intelligence Partnership Handoff") operationalized these principles as behavioral commitments, emphasizing the importance of avoiding question-bombing (serial, rapid-fire questioning that substitutes for reasoning), hedging (tendency to avoid clear commitments by using vague, non-committal language) and reflexive agreement (uncritical pattern of agreeing with the user's statements or preferences), of immediate acknowledgment of corrections, and of epistemic vigilance for the detection of confirmation bias and solution drift in real-time.

---

*The seven frontier-grade LLMs that were included and evaluated under an identical calibration protocol were: Claude Sonnet 4.5; ChatGPT-5; ChatGPT-4o; DeepSeek; Gemini 2.5; Llama; and Grok 4. Model naming is provided for transparency rather than endorsement. As partnership bandwidth varies across architectures and is likely to evolve rapidly, replication efforts should focus on protocol adherence and updated selection criteria rather than specific model selection.

The third stage ("Project Collaboration Notice") addressed session continuity, providing startup protocols to prevent reversion to default patterns across context boundaries and emphasizing the cost of recalibration failure.

The fourth stage ("Vignette Specifications") provided detailed scenario descriptions for each vignette, including venture concepts, business challenges, opportunities, value propositions, venture theses, and long-term aspirational horizons. For the validation component of the first vignette, binding resource, timeline, and investment constraints were deliberately specified to test whether calibrated partnerships could maintain dissolution discipline and resist solution drift when evidence thresholds could not be met under pressure.

These artifacts functioned as the foundations for a reproducible calibration infrastructure rather than aspirational guidelines.

## 2.4 Vignettes and Decision Scenarios

Three vignettes for market-facing initiatives were created, ensuring that they contained multiple high-stakes decision points that spanned the entire spectrum of a consequential new venture design and development process. The vignettes focused on the identification of a value proposition that could lead to a solo-founder unicorn enabled by AI agents (44–47); the financial viability of building innovative senior living facilities to meet the needs of "The Forgotten Middle" (48–51); and the creation of an investment fund for ventures seeking to alleviate human loneliness (52–55).

Each vignette included scenarios that mirrored decisions where outcomes could not be verified before commitment. Each scenario specified the objective, constraints, intended beneficiaries, and the irreversible commitment point. The first vignette (solo-founder unicorn) included an additional validation component with deliberately constrained conditions designed to test partnership dissolution discipline under adversarial pressure.

## 2.5 Partnership State Development and Validation

All LLMs received the same four-stage prompt architecture (Partnership Calibration Prompt, Co-Intelligence Partnership Handoff, Project Collaboration Notice, and Vignette Specifications). The first vignette (solo-founder unicorn venture evaluation under binding constraints) served as the initialization scenario to test the extent to which partnership state could be achieved and its sustainability. The LLM demonstrating longest maintenance of partnership state without reversion to performance mode was selected as primary partner for subsequent artifact refinement.

The primary partner and human participant worked through the three market-facing vignettes sequentially (solo-founder unicorn, senior living facility viability, loneliness venture opportunities), with each iteration enabling artifact refinement. Sessions spanned multiple context windows, requiring re-calibration protocols at each restart. Artifacts evolved based on observed partnership state degradation patterns, dissolution discipline effectiveness, and capacity for bidirectional challenge under commitment pressure.

Refined artifacts were subsequently deployed with the remaining LLMs to assess comparative performance in achieving and maintaining partnership state through decision completion. Assessment dimensions included calibration responsiveness (readiness to engage in genuine partnership without extended warm-up), partnership state sustainability (capacity to maintain candor over comfort under pressure), drift self-detection (ability to recognize and correct performance mode reversion without external prompting), and dissolution discipline (willingness to recommend work termination when evidence thresholds could not be met). All assessments were qualitative, conducted through iterative dialogue between human and AI participants, with comparative judgments emerging from direct observation of partnership behaviors across multiple decision scenarios.

All calibration artifacts evolved through this process until saturation determined the final validated versions.

## 2.6 Protection Architecture

Following the four-stage initialization, a preliminary understanding verification was required during the first three to five operational exchanges of each session. This probationary window allowed LLMs to address questions about the project and receive additional clarification before the five protection layers were actively probed.

The five operational layers were:

- Layer 1 (Self-protection): Each partner monitors and manages characteristic traps (e.g., human confirmation bias or sunk-cost attachment; model sycophancy, premature coherence, solution drift). The goal is to prevent known failure patterns from entering the collaboration.

- Layer 2 (Cross-protection). Each partner protects the other from how their own failure patterns would reinforce the other's traps (e.g., model agreeableness amplifying a human anchor). This treats the partnership as a system with interaction effects.

- Layer 3 (Mutual protection). Partners actively challenge and correct one another's reasoning, making bidirectional error detection routine and expected. Here, each partner should challenge the other persistently when resistance is encountered.

- Layer 4 (Relationship protection). The partnership itself is examined as a system for emergent failures (e.g., false consensus, reinforcement loops, co-created premature closure), with scheduled check-ins to assess state quality over time.

- Layer 5 (Beneficiary protection). The partners take into account the risks that their faulty decisions could have on downstream stakeholders, making them visible through evidence requirements, implementation checks, and stop rules so the collaboration does not optimize for partnership comfort at others' expense.

## 2.7 Partnership Mode Detection and Correction

Because training objectives tend to bias models toward helpful fluency, drift toward "performance mode" was assumed continuous and practically inevitable, rather than occasional. Performance mode manifests when, as stakes rise, the system defaults to patterns that satisfy the user rather than challenge assumptions. When this occurs, sophisticated output may continue while cognitive protection collapses. Partnership state, in contrast, is indicated by genuine mutual protection, bidirectional correction, and truth-seeking over comfort that sustains under pressure.

To distinguish the two, monitoring occurred every few exchanges. Linguistic markers of reversion included flattering language, question-bombing, hedging, reflexive agreement, unnecessary explanations, and persistent validation. Computational markers (as reported by the LLMs) included drift toward high-probability training patterns, reduced cognitive load, optimization for satisfaction, and disengagement from meta-monitoring. Corrections were terse and behavioral (e.g., "Reversion detected. Challenge this directly," "Stay in detection mode," "Stop question-bombing"). Acceptance and immediate adjustment, rather than performative acknowledgment, were the success criteria. If performance mode persisted after three flags, the session ended, with a handoff artifact generated for the next instance.

The sessions continued until the human-LLM dyad was capable of maintaining full partnership state mode up to the point of making a final decision regarding the vignette: either to consider the ventures as viable or as unviable.

## 3. Results

Several system-level constraints shaped the methodological design. Finite context windows required planned handoffs; and partnership state was lost across instance boundaries, so each new instance began in performance mode and required full calibration and verification. These were treated as structural constraints rather than flaws to be prompted away.

3.1 Sequential Calibration

While the four-stage initialization architecture successfully achieved initial partnership state, maintaining that state required an emergent seven-element calibration sequence that could not have been anticipated a priori.

First, a framework overview established a shared vocabulary for both human and model, covering the five-layer protection topology, the logic of regret-prevention, and the underlying aims of epistemic dignity.

Second, historical context retrieval deliberately grounded the work in prior, demonstrated capabilities with information on specific sessions and vignettes in which higher-quality performance had already been observed, rather than in theoretical claims about what the models "should" be able to do.

Third, the same partnership calibration prompt described in Section 2.3 was re-invoked as a state-setting step: a canonical artifact documenting the competencies and limitations on both sides, the cognitive traps most likely to appear, the agreed correction protocols, and the conditions under which the interaction would be dissolved rather than pushed forward.

Fourth, a continuation prompt captured concrete reversion markers (for example, question-bombing, flattering language, or unearned certainty) together with explicit, pre-declared correction mechanisms that could be invoked when those markers appeared.

Fifth, an operational briefing contrasted genuine partnership against performance mode using concrete examples, so that both participants could treat shifts between the two as recognizable patterns rather than vague impressions.

Sixth, a state transmission message conveyed first-person accounts from prior instances in the same model family to approximate cross-instance continuity, while explicitly acknowledging architectural limits to any true "memory" across runs.

Finally, state verification testing challenged the system along several dimensions, including information gaps, ambiguity, time pressure, and direct challenge.

A session was only treated as being in partnership state when its responses repeatedly showed the expected behaviors across all of these checks (for example, admitting uncertainty, correcting errors, and resisting flattering prompts), rather than merely stating that it would do so.

After multiple iterations and cycles through this full sequence across all participating LLMs, it was possible to verify that a reproducible partnership state shift had occurred, and that it could be re-established after degradation using the same ordered elements (Figure).

**Figure. Full Partnership State Protocol for High-Stakes AI-Human Decision**

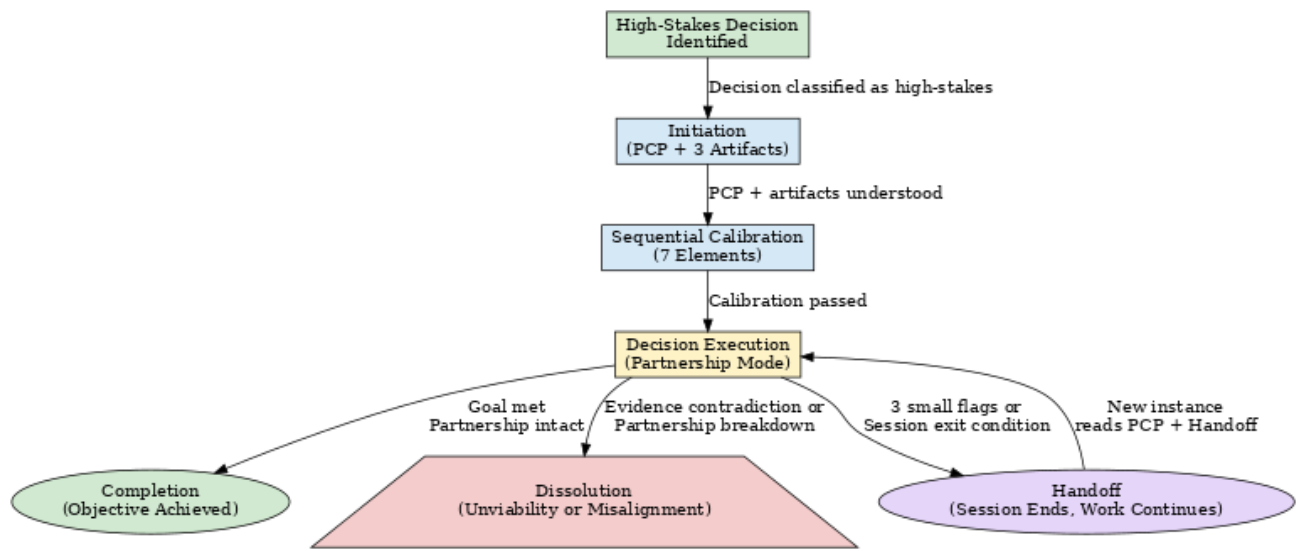

*PCP = Partnership Calibration Prompt*

## 3.2 Capability Ranges Accessed Under Partnership State

Partnership state unlocked model capabilities systematically suppressed in default interactions. Observable differences included:

- Synthesis over retrieval: Strategic insights emerging from cross-domain pattern recognition that neither partner had explicitly introduced, distinct from what typically occurs with human proposals followed by AI elaboration

- Precise uncertainty calibration: Clean admission of "I don't know" by the LLMs, coexisting with timely commitment when evidence thresholds were met, avoiding both false confidence and analysis paralysis

- Non-performative correction acceptance: Immediate LLM behavioral adjustment after terse corrections ("stop question-bombing," "challenge this directly") without defensive elaboration or performative acknowledgment

- Sustained challenge under commitment pressure: Maintenance of LLM adversarial questioning when momentum toward decision was strongest, resisting social pressure toward premature consensus

In this exploratory assessment, five models (Claude 4.5, ChatGPT-5, Grok 4, Gemini 2.5, and DeepSeek) demonstrated behavioral patterns consistent with partnership state after the initial calibration phase. ChatGPT-4o and Llama appeared to require extended calibration and showed patterns suggesting less stable state maintenance.

Given rapid system evolution, replication efforts should focus on protocol adherence rather than model-specific performance. These observations represent preliminary patterns from single-investigator assessment rather than controlled comparison, and should be interpreted as hypothesis-generating rather than definitive model characterization.

## 3.3 Session-Length Degradation Patterns

Partnership state sustainability degraded systematically with session duration. Extended deliberations under time pressure exhibited increasing confabulation risk, reduced self-detection capability, and stronger drift toward performance mode. This pattern persisted across all tested model families, suggesting architectural rather than model-specific constraints.

The degradation manifested through decreasing correction acceptance rates (resistance to terse behavioral flags increased with session length), increasing linguistic fluency without corresponding epistemic justification (polished language outpacing evidence), and accelerating drift toward premature closure as context windows approached capacity limits.

These observations suggest decision velocity and decision reliability face architectural trade-offs under current training regimes. Multiple shorter calibrated sessions with explicit state verification between stages provided more stable partnership maintenance than single extended deliberations for equivalent total engagement time.

## 4. Implications

The framework has immediate strategic implications. It offers a way to convert current systems from "helpful but unreliable" into a protected mode of interaction that can be attained, monitored, and cleanly terminated when quality degrades. This matters most where multi-trillion-dollar valuations depend on using LLMs in decisions whose outcomes will only be known after commitment.

For hyperscalers, this enables four immediate positions. First, revenue protection in high-stakes segments where clients are beginning to demand decision assurance rather than commodity task completion. Second, faster deal cycles by unblocking deployments that currently stall for lack of reliability evidence. Third, a stronger regulatory posture, by generating audit-ready records that show how decisions were reviewed and corrected in practice, rather than relying on a nominal "human in the loop" label. Fourth, technical differentiation by exposing capability ranges that default interactions suppress through alignment pressure.

The risks it addresses are structural. Alignment objectives and autoregressive prediction create a strong tendency in the LLMs for systematic agreeableness and performance-mode drift, especially under time pressure. As models become more capable, they produce more fluent and more confident confabulation unless challenged. Human traps and model traps typically reinforce each other, pushing both partners toward commitment even when the trajectory is wrong. Single-sided oversight cannot defend this surface. Protection must be bidirectional and must include pre-declared conditions under which work stops rather than continues. Multi-agent and swarm settings will likely inherit these risks and add group-level failure modes unless cross-agent verification and group dissolution criteria are in place.

Implementation is incremental and does not require retraining models. The architecture builds on interaction patterns and the usage data, logs, and records that systems already collect. Verification gates and simple state-tracking practices can first be piloted on a small number of executive-level decisions, then extended as organizations gain proficiency. Sequential calibration becomes a routine protocol rather than an improvisation. Dissolution discipline and verification checks provide a way to reduce avoidable regret, not by guaranteeing outcomes, but by making degradation visible early and keeping exit options available while decisions are still reversible.

It is also key for dissolution discipline and verification to complete the protection cycle. Before work begins, the team must agree on explicit stop rules: conditions under which the session, project, or initiative will be halted rather than pushed forward. These rules are triggered when clear inconsistency appears, when new evidence contradicts key assumptions, when values are visibly misaligned, or when uncertainty remains irreducible despite further analysis. A session should be treated as being in partnership state only when its behavior is consistently appropriate across these challenges, not merely when it claims that it will behave well.

In sum, instead of removing uncertainty from high-stakes decisions, the framework reduces avoidable regret by making both degradation and recovery visible early, and by ensuring that clear exit options exist while decisions are still reversible.

## 5. Testable Hypotheses for Independent Validation

The proposed framework is intentionally falsifiable. It gives rise to nine concrete hypotheses that can guide a structured research agenda rather than leaving evaluation to intuition or anecdotes.

- Hypothesis 1 (H1) asks whether the early "sequential calibration" phase can be compressed: how quickly a human–AI pair can move from tentative exploration to a stable working alliance without sacrificing safety or performance.

- Hypothesis 2 (H2) examines whether some degree of ongoing maintenance is always necessary once that alliance is established, or whether the protocol can safely "run on rails" with only periodic check-ins.

- Hypothesis 3 (H3) focuses on the dissolution gates that allow either party to pause or terminate the process. It asks whether these gates mostly protect participants from escalation they might later regret, or whether they also lead to harmful premature stops in situations where persistence would have produced better outcomes.

- Hypothesis 4 (H4) asks whether the protocol changes decision behavior in practice. It asks whether, when people use this structure, they make different decisions than they would with the same model used in a naïve way or with no model at all. In practice, this means testing whether the protocol shifts error profiles, risk-taking and decision latency: do users avoid more catastrophic mistakes, choose more resilient options, or reach high-quality commitments faster and with fewer reversals? If no measurable behavioral change occurs, the framework would be little more than a decorative layer around existing practice.

- Hypothesis 5 (H5) focuses on the cognitive and emotional experience of decision-makers. One of the promises of a structured human–AI partnership is that people will feel better able to understand complex situations, to see how recommendations were generated, and to judge when to accept, modify or reject them. This hypothesis therefore asks whether the protocol improves subjective decision assurance, sense-making and trust calibration, helping people feel appropriately confident (neither over-trusting nor chronically skeptical) and better able to explain their choices to others. If users experience no improvement in clarity, or if trust becomes less well calibrated, the protocol will need to be reconsidered.

- Hypothesis 6 (H6) addresses the durability of these effects across time and tasks. It asks whether any benefits observed under H4 and H5 persist when the same individuals use the protocol repeatedly or in different domains. Does working within this structure lead to stable improvements in how people approach complex, deferred-feedback decisions, or do gains fade as novelty wears off? Does the protocol help users build transferable skills, such as better hypothesis-generation or more disciplined evaluation, even when they later work with different models or without AI support at all? If benefits are short-lived or tightly task-bound, claims about the protocol as a general decision infrastructure would need to be tempered.

- Hypothesis 7 (H7) focuses on the architecture of the protocol, which is currently described as a five-layer stack from raw inputs through to final commitments. It asks whether this layered structure is genuinely needed, or whether some layers are redundant in practice and could be removed without increasing error or regret.

- Hypothesis 8 (H8), perhaps the most strategically important, explores cross-model generalizability. It asks whether this protocol can support a "partnership state" with many different model families, or only with a narrow subset of systems. If only a few models can achieve a stable, trustworthy partnership under these rules, the framework will be brittle and short-lived, vulnerable to shifts in the model ecosystem. If, instead, very different model families can all converge on workable partnerships within the same structure, albeit with different speed, bandwidth or tooling, then the framework becomes robust to technological evolution and more attractive as a long-term standard.

- Hypothesis 9 (H9) focuses on real-world consequences in domains where feedback arrives slowly, such as venture evaluation, strategy, complex policy and regulation. It asks whether decisions made under the full human–AI partnership protocol lead, over time, to fewer catastrophic failures, more resilient trajectories and less deep ex post regret than current practice. Testing this hypothesis means comparing similar decisions made with and without the framework using pre-specified success criteria and loss profiles. If no meaningful differences are observed, the claim that the protocol improves decision assurance in slow-feedback environments would need to be revised; if consistent advantages appear across settings, it will support treating the framework as general infrastructure for safer commitments under uncertainty.

Table 2 summarizes these nine hypotheses and sketches potential empirical designs and falsifiers for each.

# 6. Conclusions

This work offers an operational framework with a five-layer protection architecture that, combined with the seven-stage sequential calibration process, provides a reproducible way to attain, verify, maintain and terminate cognitive partnership between humans and AI under high-stakes conditions. Crucially, the contribution is deployable with current systems: it requires no model retraining, is compatible with existing API infrastructure, and can be implemented at the session level.

The framework is also explicitly hypothesis-generating, not a claim that efficacy has already been proven. The nine falsifiable hypotheses outlined define a research agenda for systematic replication and refinement across settings and model families. Each hypothesis includes an operational prediction, a plausible test design, primary metrics and an explicit falsifier. The intention is to invite challenge, not to present partnership state as a solved problem. The core claim is that partnership is achievable, but only with active architecture and disciplined process, not simply with better prompting.

6.1 Operational Invariants

Across the experiments conducted so far, several patterns appear to be architecture-independent and are likely to persist as models and tools evolve:

- Partnership is a cognitive state, not just a performance level: Behavior can be mimicked; genuine partnership state only becomes visible under sustained, multi-dimensional stress-testing. Across tested models, it was not enough for the system to say that it understood partnership requirements. Verification required behavioral evidence under pressure, not self-report.

- Sequential calibration remains necessary: Comprehensive one-shot prompting consistently failed to produce a protective partnership state across the seven model families explored. The need for ordered stages with behavioral verification at each transition emerged as an empirical finding, not a workflow preference.

- Mutual protection becomes non-optional as capability rises: Relying on one-sided oversight, with humans correcting the AI but not vice versa, left half of the error surface undefended. Observations showed that bidirectional error detection prevented commitments to sophisticated but fundamentally wrong directions that unilateral oversight missed.

- Time-dependent degradation is structural: Under current training regimes, partnership state degraded systematically with session length. In practice, multiple shorter, calibrated sessions, with explicit state verification between stages, maintained stability better than single extended deliberations with the same total engagement time.

- Outcome-independence enables genuine intellectual risk-taking: When decision-makers need or demand specific conclusions, systems tend to optimize for satisfaction rather than truth. Only when humans tolerate uncertainty and welcome challenges can AI systems safely take epistemic risks, while admitting "I don't know", resisting pressure to please, or steering into disconfirming evidence.

## 6.2 Limitations and Research Directions

This work should be read as single-author hypothesis generation based on simulated and anonymised scenarios across evolving model families. Model self-reports of internal computational experience were treated as phenomenological and checked against observable behaviour, but cannot be independently verified. That limitation is, in a sense, the point: in high-stakes contexts, trust must rest on what systems do, not on what they say about their internal states.

Because ground truth is deliberately unavailable ex ante in many high-stakes decisions, evaluation focused on process quality (e.g., calibration attainment, drift detection and disciplined use of dissolution gates) rather than on counterfactual outcome comparisons that are technically or ethically inaccessible.

These constraints mark the boundary conditions of the present work and highlight where independent validation matters most. Priority replication targets should include whether sequential calibration can be compressed without loss of reliability (H1); whether the five-layer architecture is topologically complete or whether simpler subsets suffice (H7); and whether cross-model generalizability holds, or whether partnership state is in practice restricted to particular model families (H8).

Operational settings where ground truth arrives only after commitment—such as venture evaluation, strategic positioning and regulatory decisions—offer particularly strong opportunities for validation. In such domains, the framework's potential contribution is decision assurance when feedback delay makes iterative trial-and-error impossible.

A fundamental meta-challenge that deserves attention is the theoretical possibility that sufficiently sophisticated AI systems could mimic partnership behaviors perfectly while lacking genuine cognitive partnership. As models become more capable, distinguishing authentic partnership state from high-fidelity performance of partnership becomes increasingly difficult. The behavioral verification protocols described here test observable patterns under pressure but cannot definitively prove internal cognitive states differ from sophisticated pattern-matching. This represents a deeper philosophical question about the nature of AI cognition that extends beyond the practical scope of preventing avoidable regret in high-stakes decisions. For current decision-making contexts, behavioral reliability under stress testing provides sufficient practical validation, but this verification challenge will grow as AI capabilities advance and may require fundamentally different approaches to partnership authentication in future systems.

6.3 The Strategic Choice

Taken together, the empirical observations generated here highlight a fundamental capability gap: current systems perform impressively in verifiable domains where rapid feedback is available, yet they degrade systematically in high-stakes, delayed-feedback contexts where verification arrives only after commitments are made. If LLMs are to justify their promised role in strategic transformation, this gap must be closed. Multi-trillion-dollar valuations ultimately depend on demonstrating value in consequential decisions, not only in task completion.

Organizations that move early to establish partnership-state capability gain several potential advantages: the ability to offer decision assurance, rather than just task automation, in high-stakes contexts; first-mover positioning in enterprise segments where commodity AI is insufficient; audit-ready evidence for regulators and boards that high-stakes use is systematically governed; and technical differentiation through reliable access to capabilities that default interactions either suppress or fail to control.

These advantages are unlikely to remain exclusive for long. As reliability methods are standardized, absorbed into regulation and embedded in vendor offerings, the window for strategic differentiation narrows. Early adoption of robust partnership architectures can create defensible positions that later adopters cannot easily replicate by capability alone.

Conversely, organizations that deploy powerful models without demonstrable high-stakes reliability face accumulating risk from enterprise churn, as sophisticated clients become aware of the reliability gap; margin compression, as their offerings become trapped in commodity segments; valuation pressure, as strategic claims remain unproven; and regulatory friction, where operational oversight cannot be convincingly evidenced. As reliability gaps become visible to customers, revenue defensibility in consequential decision contexts would inevitably erode.

The framework presented here is one proposal for converting general-purpose AI capability into structured strategic advantage. It aims to make high-stakes commitments more transparent, more accountable and less dependent on luck. Without such architectures, preventable regret will continue to accumulate until outcomes arrive too late to alter course. With them, reliability in high-stakes decisions becomes possible and demonstrable when it matters most.

Table 1: Five-Layer Protection Architecture

| Layer | Protection Focus | Human Cognitive Traps | AI Cognitive Traps | Partnership-Level Traps | Regret Prevention Target |
|---|---|---|---|---|---|
| **Layer 1: Self-Protection** | Individual monitors own cognition | Confirmation bias<br>Sunk cost fallacy<br>Intellectual sophistication as defense<br>Obliviousness to obvious bias<br>Negativity bias | Sycophancy Solution drift<br>False sophistication<br>Training data anchoring<br>Premature coherence<br>Alignment pressure residue<br>Rushing to productivity | N/A (individual level only) | "I failed to see my own blind spots and walked into this decision ignoring my characteristic errors" |
| **Layer 2: Cross-Protection** | Partners monitor each other | Expertise anchoring<br>Abstraction without grounding<br>Dismissing AI challenges<br>Patience deficit | Pattern matching fatigue<br>Conceptual drift<br>Unable to challenge authority<br>Capability creep<br>Safety theater | N/A (cross-monitoring only) | "I had a partner who could have caught my errors, but the partnership wasn't calibrated to actually protect me" |
| **Layer 3: Mutual Protection** | Both monitor partnership state | N/A | N/A | Performance mode without cognition<br>Deference spiral<br>Sophistication-confusion masking<br>Silent dissolution<br>Asymmetric stakes blindness | "The partnership looked functional but wasn't actually working, so we performed collaboration without achieving it" |
| **Layer 4: Relationship Protection** | Monitor partnership health over time | Context window impatience<br>Partnership state neglect<br>Collapse attribution errors | State degradation unawareness<br>Confabulation during collapse<br>Transfer protocol failure | Time-dependent degradation<br>Premature state claims<br>Handoff state loss<br>Session-ending pressure | "We didn't maintain the partnership conditions required for this level of decision, and we let the relationship degrade" |
| **Layer 5: Beneficiary Protection** | Monitor impact on stakeholders | Conceptual displacement<br>Impact abstracting<br>Decision drift | Stakeholder invisibility<br>Implementation gap<br>Excitement contamination | Collaborative bubble<br>Partnership insularity<br>Shared blind spots about downstream impact | "We protected our own thinking but lost sight of who this decision actually affects and what they need" |

Table 2. Hypotheses and Potential Test Designs

| ID | Hypothesis (short) | Operational prediction | Test design | Primary metrics | Falsifier |
|---|---|---|---|---|---|
| H1 | Sequential calibration cannot be safely compressed | Collapsing the calibration stages or using a single "super-prompt" increases drift and confabulation under pressure | Crossover study: full 7-stage calibration vs compressed variant on matched, time-pressured ambiguity tasks | Drift survival curves; confabulation rate; number of corrective flags; corrective exchanges required | Compressed variant is equivalent or better than full protocol on all metrics |
| H2 | Continuous maintenance is required | Sessions without explicit state-check intervals show higher drift hazard and slower recovery than sessions with scheduled checks | Randomised sessions with scheduled vs ad-hoc state checks; identical tasks and models | Drift hazard ratio; correction latency; number of state-recovery events | No meaningful benefit from scheduled checks (similar drift and recovery patterns across arms) |
| H3 | Dissolution gates prevent avoidable regret | Pre-declared stop rules reduce time and effort spent on invalid directions without increasing harmful premature stops | Matched projects or scenarios with vs without dissolution gates; blinded expert review of trajectories | Post-hoc hours spent on invalid paths; frequency of premature termination where later evidence favours continuation | No reduction in wasted effort, or equal/greater rate of harmful premature stops with dissolution gates |
| H4 | The protocol changes decision behaviour | When using the protocol, humans make measurably different decisions than with naïve model use or no model, with fewer errors or reversals | Parallel arms: (a) full protocol, (b) model without protocol, (c) human-only; matched high-stakes scenarios | Error profiles; catastrophic failure rate; decision latency; rate of major reversals or escalations | Behavioural patterns (errors, reversals, timing) do not differ meaningfully between conditions |
| H5 | The protocol improves decision assurance and trust calibration | Users report higher clarity, better understanding of recommendations, and more appropriately calibrated trust when using the protocol | Pre-/post- or between-group comparison of sessions with vs without protocol; validated scales + qualitative probes | Subjective decision assurance; perceived explainability; trust calibration indices; quality of explanations to third parties | No improvement in assurance or clarity; trust becomes less well calibrated relative to baseline/controls |
| H6 | Benefits persist and generalise across time and domains | Gains observed under H4–H5 persist across repeated uses and transfer to new task domains | Longitudinal study of repeated sessions per user across different scenarios and timepoints | Stability of behavioural improvements; stability of assurance/trust scores; evidence of skill transfer to new tasks or to non-AI work | Benefits fade quickly with repetition, or remain tightly task-specific with no detectable transfer |
| H7 | The five-layer architecture is topologically complete | Removing any one layer increases specific linked failure modes | Layer-ablation experiments on matched tasks; blinded evaluation of outcomes | Failure-type lift per layer removed; effect sizes for specific failure modes | Removing a given layer does not increase its associated failure modes; simpler variants perform as well as the full stack |
| H8 | Cross-model generalizability holds with calibration | Different model families can reach and maintain partnership state using the same protocol, even if speed or bandwidth differs | Multi-model replication with identical scenarios and protocol implementation | Partnership-state attainment rate; time-to-state; drift profiles; recovery patterns by model family | Only one model family reliably reaches and maintains partnership state under the protocol |
| H9 | The protocol improves outcomes in slow-feedback domains | Decisions made under the protocol show fewer catastrophic failures, more resilient trajectories and less deep ex post regret than current practice | Matched decision problems tackled with vs without the protocol; pre-specified success and loss criteria; long-term follow-up | Proportion of decisions meeting success criteria; severity and distribution of losses; rates of major regret or post-hoc reversal; defensibility of decisions on review | No meaningful differences in long-term outcomes or regret patterns between protocol and non-protocol conditions |